\documentclass[10pt,journal,compsoc]{IEEEtran}

\ifCLASSOPTIONcompsoc
  \usepackage[nocompress]{cite}
\else
  \usepackage{cite}
\fi

\ifCLASSINFOpdf
   \usepackage[pdftex]{graphicx}
   \graphicspath{{./fig/}}
\else
\fi

\usepackage{amsmath}
\usepackage{amssymb}
\usepackage{epsfig}
\usepackage{graphicx}
\usepackage{bm}
\usepackage{color}
\usepackage{url}

\DeclareMathOperator*{\argmin}{argmin}

\begin{document}
%
\title{Audio-driven Talking Face Video Generation with Learning-based Personalized Head Pose}

\author{Ran Yi, Zipeng Ye,
        Juyong~Zhang,~\IEEEmembership{Member,~IEEE,}
        Hujun~Bao,~\IEEEmembership{Member,~IEEE,}
        and \\Yong-Jin~Liu,~\IEEEmembership{Senior Member,~IEEE}
\IEEEcompsocitemizethanks{\IEEEcompsocthanksitem R. Yi, Z. Ye, Y.-J. Liu are with MOE-Key Laboratory of Pervasive Computing, the Department of Computer Science and Technology, Tsinghua University, Beijing, China. Y.-J. Liu is the corresponding author. E-mail: liuyongjin@tsinghua.edu.cn.
\IEEEcompsocthanksitem J. Zhang is with the School of Mathematical
Sciences, University of Science and Technology of China, Hefei, China.
\IEEEcompsocthanksitem H. Bao is with the college of Computer Science and Technology, Zhejiang University, Hangzhou, China.}
}

\markboth{}%
{Yi \MakeLowercase{\textit{et al.}}: Audio-driven Talking Face Video Generation with Learning-based Personalized Head Pose}

\IEEEtitleabstractindextext{%
\begin{abstract}
Real-world talking faces often accompany with natural head movement. However, most existing talking face video generation methods only consider facial animation with fixed head pose. In this paper, we address this problem by proposing a deep neural network model that takes an audio signal A of a source person and a very short video V of a target person as input, and outputs a synthesized high-quality talking face video with personalized head pose (making use of the visual information in V), expression and lip synchronization (by considering both A and V). The most challenging issue in our work is that natural poses often cause in-plane and out-of-plane head rotations, which makes synthesized talking face video far from realistic. To address this challenge, we reconstruct 3D face animation and re-render it into synthesized frames. To fine tune these frames into realistic ones with smooth
background transition, we propose a novel memory-augmented GAN module. By first training a general mapping based on a publicly available dataset and fine-tuning the mapping using the input short video of target person, we develop an effective strategy that only requires a small number of frames (about 300 frames) to learn personalized talking behavior including head pose. Extensive experiments and two user studies show that our method can generate high-quality (i.e., personalized head movements, expressions and good lip synchronization) talking face videos, which are naturally looking with more distinguishing head movement effects than the state-of-the-art methods.
\end{abstract}

}

\maketitle

\IEEEdisplaynontitleabstractindextext

\IEEEpeerreviewmaketitle

\section{Introduction}

Visual and auditory modalities are two important sensory channels in human-to-human or human-to-machine interaction. The information in these two modalities are strongly correlated \cite{Nazzaro70}. Recently, cross-modality learning and modeling have  attracted more and more attention in interdisciplinary research, including computer vision, computer graphics and multimedia (e.g., \cite{ChenMDX19,ChungZ16,ChungJZ17,Freeman14,OhDKMFRM19,SuwajanakornSK17}).

In this paper, we focus on talking face video generation that transfers a segment of audio signal of a source person into the visual information of a target person. This kind of audio-driven vision models have a wide range of applications, such as bandwidth-limited video transformation, virtual anchors and role-playing game/move generation, etc. Recently, many works have been proposed for this purpose (e.g., \cite{ChenMDX19,ChungJZ17,SongZLWQ19,ZhouLLLW19}). However, most of them only consider facial animation with {\it fixed} head pose.

In real-world scenarios, natural head movement plays an important role in high-quality communication \cite{GlowinskiDCVMS11} and human perception is very sensitive to subtle head movement in real videos. In fact, human can easily feel uncomfortable in communication by talking with fixed head pose. In this paper, we propose a deep neural network model to generate an audio-driven high-quality talking face video with personalized head pose.

Inferring head pose from speech (abbreviated as pose-from-speech) is not a new idea (e.g., \cite{GreenwoodLM17,GreenwoodML18}). Although some measurable correlations have been observed between speech and head pose \cite{BussoDGNN07,OhDKMFRM19}, predicting head motion from speech is still a challenging problem. A practical method was suggested in \cite{GreenwoodML18} that first infers facial activity from speech and then models head pose from facial features. In our work, by observing that simultaneously learning two related tasks in deep network
may help improve the performance of both tasks, we simultaneously infers facial expressions and head pose from speech.

Since natural head poses often cause in-plane and out-of-plane head rotations, it is very challenging to synthesize a realistic talking face video with high-quality facial animation and smooth background transition. To circumvent the difficult pose-from-speech problem and focus on addressing the realistic video synthesis challenge, we design the input of our system to include a segment of audio signal of a source person and a short (only a few seconds) talking face video of a target person. Note that with the popularization of smartphone, the cost of capturing a very short video (e.g., 10 seconds) is almost the same as taking a photo (e.g., selfie). Therefore we use both facial and audio information in the input short video to learn the personalized talking behavior of the target person (e.g., lip and head movements), which greatly simplifies our system.

To output a high-quality synthesized video of the target person with personalized head pose when speaking the input audio signal of source person, our system reconstructs 3D face animation and re-renders it into video frames. Given a light-weight rendering engine with limited information, these rendered frames are often far from realistic. We then propose a novel memory-augmented GAN module that can refine the rough rendered frames into realistic frames with smooth transition, according to the identity feature of the target person. To the best of our knowledge, our proposed method is the first system that can transfer the audio signal of an {\it arbitrary} source person into the face talking video of an {\it arbitrary} target person with {\it personalized} head pose. As a comparison, the previous work \cite{SuwajanakornSK17}
can only generate a high-quality talking face video with personalized head pose for a {\it specified} person (i.e., Obama) --- since it requires a large number of training data related to this specified person --- and thus, it cannot generalize to arbitrary subjects. Furthermore, when the input short talking face video is not available, our method can also use a face image as input and achieves comparable lip synchronization and video quality with previous methods \cite{ChenMDX19,ChungJZ17,ZhouLLLW19}.
Our code is publicly available\footnote{\url{https://github.com/yiranran/Audio-driven-TalkingFace-HeadPose}}.

The contributions of this paper are mainly three-fold:
\begin{itemize}
\item We propose a novel deep neural network model that can transfer an audio signal of arbitrary source person into a high-quality talking face video of arbitrary target person, with {\it personalized} head pose and lip synchronization.
\item Different from the network \cite{KimCTXTNPRZT18} that fine tunes the rendering of a {\it specified} parametric face model into photo-realistic video frames, our memory-augmented GAN module can generate photo-realistic video frames for {\it various} face identities (i.e., corresponding to different target person).
\item By first training a general mapping based on a publicly available dataset \cite{ChungZ16} and fine-tuning the mapping using the input short video of the target person, we develop an effective strategy that only requires a small number of frames (about 300 frames) to learn personalized talking behavior including head pose.
\end{itemize}

\begin{figure*}[t]
\centering
\includegraphics[width = 1.0\textwidth]{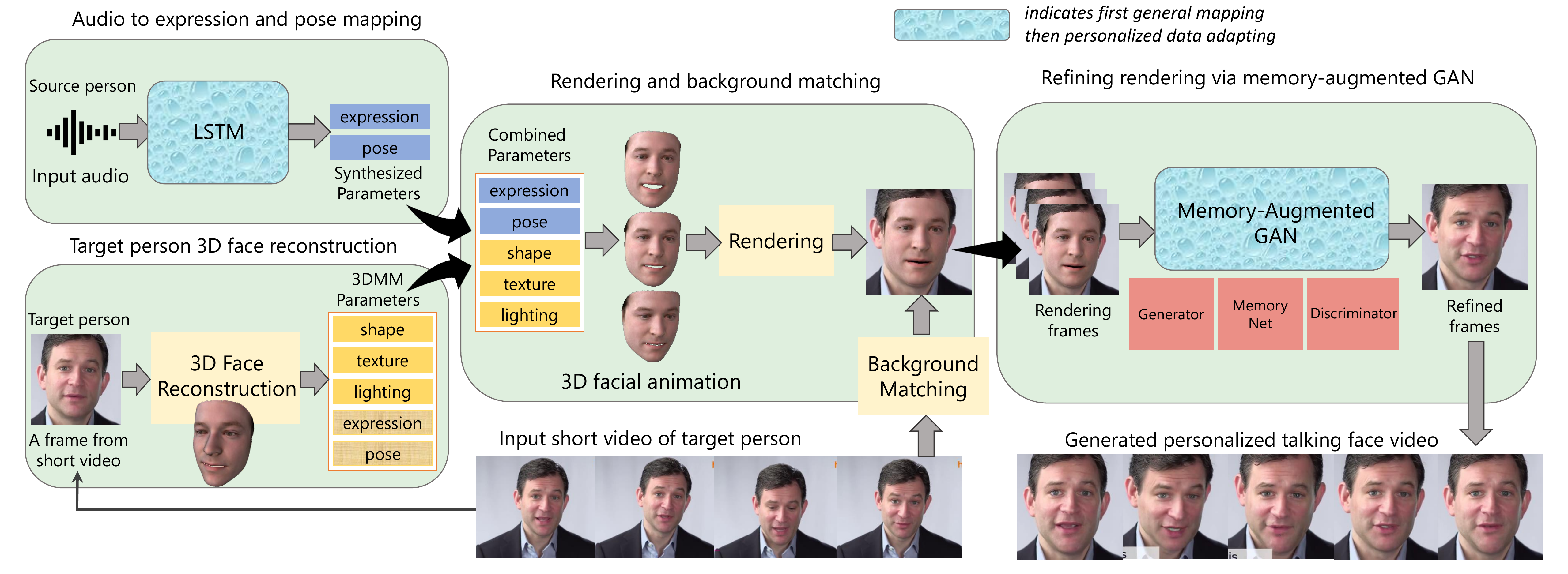}
\caption{Flowchart of our method. (Stage 1) We train a general mapping from the input audio to the facial expression and common head pose. Then, we reconstruct the 3D face and fine
tune the general mapping to learn personalized talking behavior from the input video. So we can obtain the 3D facial animation with personalized head pose. (Stage 2) We render the 3D facial animation into video frames using the texture and lighting information obtained from input video. Then we fine tune these synthesized frames into realistic frames using a novel memory-augmented GAN module.
}
\label{fig:pipeline}
\end{figure*}

\section{Related Work}

\subsection{Talking face generation}

Existing talking face video generation methods can be broadly categorised into two classes according to the driven signal. One driven signal is video frames~\cite{ThiesZSTN16,Averbuch-ElorCK17,PumarolaAMSM18,WilesKZ18,KimCTXTNPRZT18,ZakharovSBL19,ZhangZHLLL19} and the other is audio~\cite{FanWSX15,SuwajanakornSK17,WilesKZ18,ChungJZ17,ZhouLLLW19,ChenMDX19,SongZLWQ19,VougioukasPP19,Thies2019arxiv}.
Video-driven talking face video generation (a.k.a face reenactment) transferred expression and sometimes head pose from a driving frame to a face image of target actor.
Traditional optimization methods transferred expression using 3DMM parameters~\cite{ThiesZSTN16,ThiesZSTN18} or image warping~\cite{Averbuch-ElorCK17}.
Learning-based methods~\cite{KimCTXTNPRZT18,ZakharovSBL19} were trained by videos of target actor or general audio-visual data using GAN model conditioned on image or additional landmarks. Video-frame-driven methods only use one modality, i.e., visual information.

Audio-driven methods make use of both visual and auditory modalities, which can be further classified into two sub-classes: talking face video generation for specific face \cite{FanWSX15,SuwajanakornSK17} and for arbitrary target face \cite{ChenMDX19,ChungJZ17,SongZLWQ19,ZhouLLLW19}. The latter methods usually take a clip of audio and one arbitrary face image as input.
Chung et al.~\cite{ChungJZ17} learned a joint embedding of the face and audio signal, and used an encoder-decoder CNN model to generate talking face video.
Zhou et al.~\cite{ZhouLLLW19} proposed a method in which both audio and video can serve as input by learning joint audio-visual representation.
Chen et al.~\cite{ChenMDX19} first transferred the audio to facial landmarks and then generated video frames conditioned on the landmarks.
Song et al.~\cite{SongZLWQ19} proposed a conditional recurrent adversarial network that integrated audio and image features in recurrent units.
However, in talking face videos generated by these 2D methods, the head pose is almost fixed during talking.
This drawback is caused by the defect inherent in 2D-based methods, since it is difficult to only use 2D information alone for naturally modeling the change of pose.
Although Song et al.~\cite{SongZLWQ19} mentioned that their method can be extended to personalized pose for a special case, full details on this extension were not yet presented.
In comparison, we introduce 3D geometry information into the proposed system to simultaneously model personalized head pose, expression and lip synchronization.

\subsection{3D face reconstruction}

3D face reconstruction aims to reconstruct 3D shape and appearance of human face from 2D images. A large number of methods have been proposed in this area and the reader is referred to the survey~\cite{zollhofer2018state} and reference therein.
Most of these methods were based on 3D Morphable Model (3DMM)~\cite{BlanzV99}, which learned a PCA basis from scanned 3D face data set to represent general face shapes.
Traditional methods fit 3DMM by an analysis-by-synthesis approach, which optimized 3DMM parameters by minimizing difference between rendered reconstruction and the given image~\cite{BlanzV99,GarridoZCVVPT16,JiangZDLL18}.

Learning-based methods~\cite{RichardsonSK16,ZhuLLSL16,RichardsonSOK17,JacksonBAT17,SelaRK17,TewariZK0BPT17,TranL18,GenovaCMSVF18,GecerPKZ19,Guo20193DFace,Deng2019accurate} used CNN to learn a mapping from face images to 3DMM parameters.
To deal with the lack of sufficient training data, some methods used synthetic data~\cite{ZhuLLSL16,RichardsonSK16,SelaRK17,Guo20193DFace} while others use unsupervised or weakly-supervised learning~\cite{TewariZK0BPT17,TranL18,GenovaCMSVF18,Deng2019accurate}.
In this paper, we adopt the method ~\cite{Deng2019accurate} for 3D face reconstruction.

\subsection{GANs and memory networks.}

Generative Adversarial Networks (GANs)~\cite{GoodfellowPMXWOCB14} have been successfully applied to many computer vision problems.
The Pix2Pix proposed by Isola et al. \cite{IsolaZZE17} has shown great power in image-to-image translation between two different domains.
Later it was extended to video-to-video synthesis~\cite{Wang0ZYTKC18,wang2019fewshotvid2vid}.
It has also been applied to the field of facial animation and texture synthesis.
Kim et al.~\cite{KimCTXTNPRZT18} use a GAN to transform rendered face image to realistic video frame. Although this method can achieve good results, it was only suitable for a specific target person, and it had to be trained by thousands of samples related to this specific person.
Olszewski et al.~\cite{OlszewskiLYZYHX17} proposed a network to generate realistic dynamic textures.

Memory network is a scheme to augment neural networks using external memory. It has been applied to question-answering systems~\cite{SukhbaatarSWF15,KumarIOIBGZPS16}, summarization~\cite{KimKK19}, image captioning~\cite{ParkKK17}, and image colorization~\cite{YooBCLCC19}.
Since this scheme can remember selected critical information, it is effective for one-shot or few-shot learning.
In this paper, we use a GAN augmented with memory networks to fine tune rendered frames into  realistic frames for arbitrary person.

\section{Our Method}

In this paper, we tackle the problem of generating high-quality talking face video, when given an audio speech of a source person and a short video (about 10 seconds) of a target person. In addition to learn the transformation from the audio speech to lip motion and face expression, our talking face generation also considers the personalized talking behavior (i.e., head pose) of the target person.

To achieve this goal, our idea is to use {\it 3D facial animation with personalized head pose} as the kernel to bridge the gap between audio-visual-driven head pose learning and realistic talking face video generation. The flowchart of our method is illustrated in Figure \ref{fig:pipeline}, which can be interpreted in the following two stages.

{\it Stage 1: from audio-visual information to 3D facial animation.}
We use the LRW video dataset \cite{ChungZ16} to train a general mapping from the audio speech to the facial expression and common head pose. Then, given an audio signal and a short video, we first reconstruct the 3D face (Section \ref{ssec:facerecon}) and fine tune the general mapping to learn personalized talking behavior from the input video (Section \ref{ssec:audio}). To this end, we obtain the 3D facial animation with personalized head pose.

{\it Stage 2: from 3D facial animation to realistic talking face video generation.}
We render the 3D facial animation into video frames using the texture and lighting information obtained from input video. With these limited information, the graphic engine can only provide a rough rendering effect that is usually not realistic enough for a high-quality video. To refine these synthesized frames into realistic ones, we propose a novel memory-augmented GAN module (Section \ref{ssec:gan}) that was also trained by the LRW video dataset.
This GAN module can deal with various identities and generate high-quality frames containing realistic talking faces that matches the face identity extracted from input video.

Note that both mapping in the above two stages involves two steps: one step is the general mapping learned from the LRW video dataset and the second is a light-weight fine-tuning step that learns/retrieves personalized talking or rendering information from the input video.

\subsection{3D face reconstruction}
\label{ssec:facerecon}

We adopt a state-of-the-art deep learning based method~\cite{Deng2019accurate} for 3D face reconstruction.
It uses a CNN to fit a parametric model of 3D face geometry, texture and illumination to an input face photo $\mathbf{I}$.
This method reconstructs the 3DMM coefficients $\chi(\mathbf{I})=\{\boldsymbol{\alpha},\boldsymbol{\beta},\boldsymbol{\delta},\boldsymbol{\gamma},\mathbf{p}\} \in \mathbb{R}^{257}$, where $\boldsymbol{\alpha} \in \mathbb{R}^{80}$ is the coefficient vector for face identity, $\boldsymbol{\beta} \in \mathbb{R}^{64}$ is for expression, $\delta \in \mathbb{R}^{80}$ is for texture, $\boldsymbol{\gamma} \in \mathbb{R}^{27}$ is the coefficient vector for illumination, and $\mathbf{p} \in \mathbb{R}^{6}$ is the pose vector including rotation and translation.
Then the face shape $S$ and face texture $T$ can be represented as $\mathbf{S}=\mathbf{\bar{S}} + \mathbf{B}_{id} \boldsymbol\alpha + \mathbf{B}_{exp} \boldsymbol\beta$, $\mathbf{T}=\mathbf{\bar{T}}+\mathbf{B}_{tex} \boldsymbol\delta$, where $\mathbf{\bar{S}}$ and $\mathbf{\bar{T}}$ are average shape and texture, $\mathbf{B}_{id}$, $\mathbf{B}_{exp}$ and $\mathbf{B}_{tex}$ are PCA basis for shape, expression and texture separately. Basel Face Model~\cite{PaysanKARV09} is used for $\mathbf{B}_{id}$ and $\mathbf{B}_{tex}$, and FaceWareHouse~\cite{CaoWZTZ14} is used for $\mathbf{B}_{exp}$.

The illumination is computed using the Lambertian surface assumption and approximated with spherical harmonics (SH) basis functions~\cite{RamamoorthiH01a}. The irradiance of vertex $v_i$ with normal vector $\mathbf{n}_i$ and texture $\mathbf{t}_i$ is $C(\mathbf{n}_i,\mathbf{t}_i,\boldsymbol\gamma) = \mathbf{t}_i \sum_{b=1}^{B^2}\gamma_b \Phi_b(\mathbf{n}_i)$, where $\Phi_b: \mathbb{R}^3 \rightarrow \mathbb{R}$ are SH basis functions, $\gamma_b$ are SH coefficients and $B=3$ is the number of SH bands. The pose is represented by rotation angles and translation. A perspective camera model is used to project the 3D face model onto the image plane.

\subsection{Mapping from audio to expression and pose}
\label{ssec:audio}

It is well recognized that the audio signal has strong correlation with lip and lower-half face movements.
However, talking faces with only lower-half face movements are stiff and far from natural.
In other words, upper-half face (including eyes and brows) movements and head pose are also essential for a natural talking face.
We use both the audio information and the 3D face geometry information extracted from input video to establish a mapping from the input audio to the facial expression and head pose.
Note that although a person may have different head poses when speaking the same word, the speaking style in a short period is often consistent and we provide a correlation analysis between audio and pose in Appendix A.

We extract the Mel-frequency cepstral coefficients (MFCC) feature of the input audio, and model the facial expression and head pose using 3DMM coefficients. To establish the mapping inbetween, we design a LSTM network as follows.
Given the MFCC features of an audio sequence
$\mathbf{s}=\{s^{(1)},\ldots,s^{(T)}\}$,
a ground-truth expression coefficient sequence
$\boldsymbol{\beta}=\{\beta^{(1)},\ldots,\beta^{(T)}\}$,
and a ground-truth pose vector sequence
$\mathbf{p}=\{p^{(1)},\ldots,p^{(T)}\}$,
we generate predicted expression coefficient sequence
$\widetilde{\boldsymbol\beta}=\{\widetilde\beta^{(1)},\ldots,\widetilde\beta^{(T)}\}$ and pose vector sequence $\widetilde{\mathbf{p}}=\{\widetilde{p}^{(1)},\ldots,\widetilde{p}^{(T)}\}$.
Denoting the LSTM network as $R$, our audio-to-expression-and-head pose mapping can be formulated as
\begin{equation}
[\widetilde\beta^{(t)},\widetilde{p}^{(t)},h^{(t)},c^{(t)}] = R(E(s^{(t)}),h^{(t-1)},c^{(t-1)}),
\label{eq:lstm}
\end{equation}
where $E$ is an additional audio encoder that is applied to the MFCC feature of audio sequences $s^{(t)}$, and $h^{(t)},c^{(t)}$ are hidden state and cell state of LSTM unit at time $t$ respectively.

We use a loss function containing four loss terms to optimize the network: a mean squared error (MSE) loss for expression coefficients, a MSE loss for pose coefficients, an inter-frame continuity loss for pose, and an inter-frame continuity loss for expression.
Denote the shorthand notation of Eq.~\eqref{eq:lstm} as $\widetilde{\boldsymbol\beta}=\phi_1(\mathbf{s}), \widetilde{\mathbf{p}}=\phi_2(\mathbf{s})$, the loss function is formulated as:
\begin{equation}
\begin{aligned}
\mathcal{L}(R,E) &= \mathbb{E}_{\mathbf{s},\boldsymbol{\beta}}[(\boldsymbol{\beta}-\phi_1(\mathbf{s}))^2] + \lambda_1  \mathbb{E}_{\mathbf{s},\mathbf{p}}[(\mathbf{p}-\phi_2(\mathbf{s}))^2] \\
&+ \lambda_2 \mathbb{E}_{\mathbf{s}}[\sum_{t=0}^{T-1}(\phi_2(\mathbf{s})^{(t+1)}-\phi_2(\mathbf{s})^{(t)})^2] \\
&+ \lambda_3 \mathbb{E}_{\mathbf{s}}[\sum_{t=0}^{T-1}(\phi_1(\mathbf{s})^{(t+1)}-\phi_1(\mathbf{s})^{(t)})^2],
\end{aligned}
\label{eq:lstmloss}
\end{equation}
where inter-frame continuity loss is computed by the squared $L_2$ norm of the gradient of the pose / expression.

\subsection{Rendering and background matching}

\subsubsection{Rendering}

By reconstructing the 3D face of the target person (Section~\ref{ssec:facerecon}) and generating the expression and pose sequences (Section~\ref{ssec:audio}), we collect a mixed sequence of 3DMM coefficients synchronized with audio speech, in which the identity, texture and illumination coefficients are from the target person, and expression and pose coefficients are from the audio.
Given this mixed sequence of 3DMM coefficients, we can render a face image sequence using the rendering engine in \cite{GenovaCMSVF18}.

If we compute the albedos from reconstructed 3DMM coefficients, these albedos are of low-frequency and too smooth, resulting in the rendered face images that do not appear visually similar to the input face images. An alternative is to compute a {\it detailed} albedo from input face images. I.e., we first project the reconstructed 3D shape (a face mesh) onto the image plane, and then we assign the pixel color to each mesh vertex. In this way, the albedo is computed by dividing illumination. Finally, the albedo from the frame with the most neutral expression and the smallest rotation angles is set as the albedo of the video.

We use the above mentioned both schemes in our method.
In the general mapping, we use the detailed albedo for rendering, because videos in the LRW dataset are very short (about 1 second).
In the personalized mapping (i.e., tuning by input short video), we use the low-frequency albedo to tolerate the change of head pose, and the input video (about 10 seconds) can provide more training data of the target person to fine tune the synthesized frames (rendered with a low-frequency albedo) into realistic ones.


\begin{figure*}[t]
\centering
\includegraphics[width = 1.0\textwidth]{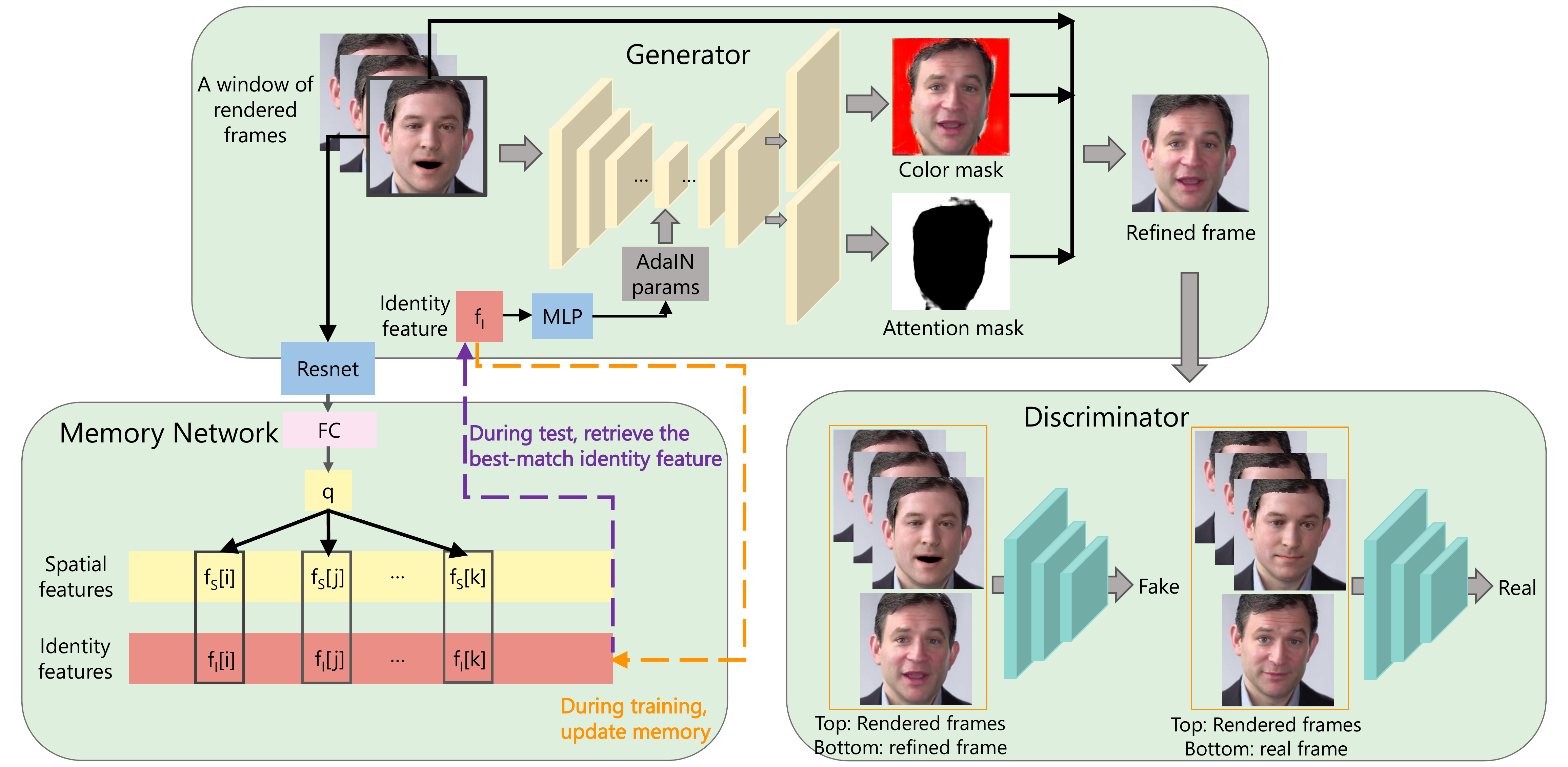}
\caption{Our memory-augmented GAN for refining rendered frames into realistic frames. The generator takes a window of rendering frames and an identity feature as input, and generate a refined frame based on attention mechanism. Discriminator judges whether a frame is real or not. The memory network is introduced to remember representative identities during training and retrieve the best-match identity feature during test. During the training, the memory network is updated by paired spatial features and ground-truth identity features. During the test, the memory network retrieves the best-match identity feature using the spatial feature as query.}
\label{fig:gan}
\end{figure*}

\subsubsection{Background matching}

So far the rendered frames only have the facial part, without the hair and background regions that are also essential for a realistic talking face video.
An intuitive solution is to match a background from the input video by matching the head pose.
However, for a short video of about 10 seconds, we only have less than 300 frames to select a suitable background, which is very few and can be regarded as very sparse points in the possible high-dimensional pose space. Our experiment also shows that this intuitive solution cannot produce good video frames.

In our method, we propose to extract some keyframes from the synthesized pose sequence, where the keyframes correspond to critical head movements in the synthesized pose sequence. We choose the key frames to be the frames with largest head orientation in one axis in a short period of time, e.g., the frame with leftmost or rightmost head pose.
Then we only match backgrounds for these keyframes. We call these matched backgrounds as key backgrounds.
For those frames between two neighboring keyframes, we use linear interpolation to determine their backgrounds. The pose in each frame is also modified to fit the background.
Finally the whole rendered frames are assembled by including the matched backgrounds.

If only a signal face image $I$ is input instead of a short video, we obtain the matched background by rotating $I$ to the predicted pose using the face profiling method in~\cite{ZhuLLSL16}.

\subsection{Memory-augmented GAN for refining frames}
\label{ssec:gan}

The synthesized frames rendered by the light-weight graphic engine \cite{GenovaCMSVF18} are usually far from realistic. To refine these frames into realistic ones, we propose a memory-augmented GAN. The differences between our method and the previous GAN-based face reenactment (FR) methods~\cite{KimCTXTNPRZT18} are:
\begin{itemize}
\item FR only refines the frames for a single, specified face identity, while our method can deal with various face identities. I.e.,
given different identity features of target faces, our method can output different frame refinement effects with the same GAN model.
\item FR uses thousands of frames to train a network for a single, specified face identity, while we only use a few frames for each identity in the general mapping learning.
Based on the general mapping, we fine tune the network using a small number of frames for the target face (from the input short video).
\end{itemize}

We model the frame refinement process as a function $\Phi$ that maps from the rendered frame (i.e., synthesized frame rendered by the graphic engine) domain $\mathcal{R}$ to the real frame domain $\mathcal{T}$ using paired training data $\{(r_i,g_i)\}$, $r_i\in\mathcal{R}$ and $g_i\in\mathcal{T}$.
To handle multiple-identity refinement, we build a GAN network that consists of a conditional generator $G$, a conditional discriminator $D$ and an additional memory network $M$ (Figure \ref{fig:gan}).
The memory network stores paired features, i.e., (spatial feature, identity feature), which are updated during the training process. Its role is to remember representative identities including rare instances in the training set, and retrieve the best-match identity feature during the test.
The conditional generator takes a window of rendered frames (i.e., a subset of 3 adjacent frames $r_{t-2},r_{t-1},r_t$) and an identity feature as input, and synthesize a refined frame $o_t$ using the U-Net~\cite{RonnebergerFB15} with AdaIN~\cite{HuangB17}.
The conditional discriminator takes a window of rendered frames and either a refined frame or a real frame as input, and decides whether the frame is real or synthesized.

{\bf Attention-based generator $G$.}
We use an attention-based generator to refine rendering frames.
Given a window of rendered frames $(r_{t-2},r_{t-1},r_t)$ and an identity feature $f_t$ (extracted from ArcFace~\cite{DengGXZ19}), the generator synthesizes both a color mask $C_t$ and an attention mask $A_t$, and outputs a refined frame $o_t$ that is the weighted average of the rendered frame and color mask:
\begin{equation}
o_t = A_t \cdot r_t + (1-A_t) \cdot C_t
\end{equation}
The attention mask specifies how much each pixel in the generated color mask contributes to the final refinement.
Our generator architecture is based on a U-Net structure\footnote{The attention mechanism has also been used in the work~\cite{PumarolaAMSM18}. However, the difference in our network is that we also input an additional identity feature into the network, which enables generating different refining effects for different identities.} and has two modifications. (1) To generate two outputs (i.e., color and attention masks), we modify the last convolution block to two parallel convolution blocks, in which each one generates one mask.
(2) To take both a window of rendered frames and an identity feature as input, we adopt AdaIN~\cite{HuangB17} to incorporate identity features into our network, where AdaIN parameters are generated from input identity features. Experimental results show that our network can generate delicate target-person-dependent texture for various identities.

\begin{figure*}[t]
\centering
\includegraphics[width = 1.0\textwidth]{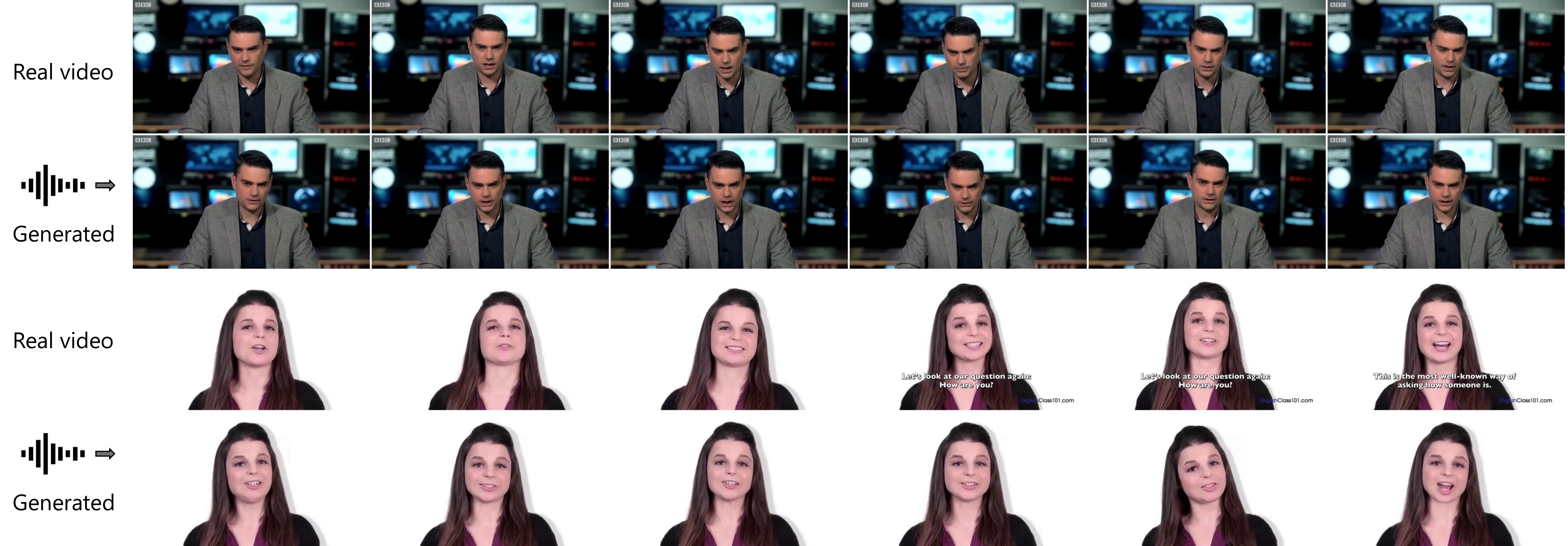}
\caption{Comparison of real videos with natural head pose and our generated talking face videos with personalized behavior. Our method can achieve both good lip synchronization and personalized head pose.} 
\label{fig:personalized_rst}
\end{figure*}

{\bf Memory network $M$.}
We use a memory network to remember representative identities including rare instances in the training set, so that during the test we can retrieve similar identity feature from it.
We adapt the network in \cite{YooBCLCC19} in our system by modifying it to output continuous frames.
In particular, our memory network stores paired {\it spatial features} and {\it identity features}.
The spatial feature is extracted by (1) feeding the input rendered frame into ResNet18~\cite{HeZRS16} pre-trained on ImageNet~\cite{DengDSLL009}, (2) extracting the `pool5' feature, and (3) passing the `pool5' feature to a learn-able fully connected layer and normalization. The paired identity feature is extracted by feeding the corresponding ground-truth frame into ArcFace~\cite{DengGXZ19}.

During the training, we update the memory network using paired features extracted from the training set. This updating includes (1) a threshold triplet loss\footnote{We use the cosine similarity for both spatial and identity features.} \cite{YooBCLCC19} to make spatial features of similar identities closer and spatial features of different identities farther, and (2) a memory item updating process, where either an existing feature pair is updated or an old pair is replaced\footnote{An old pair is replaced when the similarity between current identity feature and the closest identity feature is smaller than a threshold.} by a new pair. During the test, we retrieve the identity feature by using the spatial feature as query, finding its nearest spatial feature in memory and returning the corresponding identity feature. Noting that directly feeding this feature into the generator may lead to jittering effects, we smooth the retrieved features in multiple adjacent frames by interpolation and use the smoothed features as inputs for the generator.

{\bf Discriminator $D$.}
The conditional discriminator takes a window of rendered frames and a checking frame (either a refined frame or a real frame) as input, and discriminates whether the checking frame is real or not. We adopt PatchGAN~\cite{IsolaZZE17} architecture as our discriminator.

{\bf Loss function.}
The loss function of our GAN model\footnote{Note that during the training process, the memory network is updated separatedly and GAN is trained after each updating of the memory network.} has three terms: a GAN loss, an $L_1$ loss, and an attention loss~\cite{PumarolaAMSM18} to prevent the attention mask $A$ from saturation, which also enforces the smoothness of the attention mask. Denoting the input rendered frames as $r$, the identity feature as $f$, and the ground truth real frames as $g$, the loss function is formulated as:
\begin{equation}
\begin{aligned}
\mathcal{L}(G,D) &= (\mathbb{E}_{r,g}[\log D(r,g)] + \mathbb{E}_{r}[\log (1-D(r,G(r,f)))]) \\
&+ \lambda_1 \mathbb{E}_{r,g}[||g - G(r,f)||_1] + \lambda_2 \mathbb{E}_{r}[||A||_2]\\
&+ \lambda_3 \mathbb{E}_{r}[\sum_{i,j}^{H,W}(A_{i+1,j}-A_{i,j})^2 + (A_{i,j+1}-A_{i,j})^2]
\end{aligned}
\label{eq:ganloss}
\end{equation}
We train the GAN model to optimize the loss function:
\begin{equation}
\begin{aligned}
G^* = \argmin_{G} \max_{D} \mathcal{L}(G,D)
\end{aligned}
\end{equation}

\section{Experiments}

We implemented our method in PyTorch.
All experiments are performed on a PC with a Titan Xp GPU.
The code is publicly available\footnote{\url{https://github.com/yiranran/Audio-driven-TalkingFace-HeadPose}}.
The dynamic results in this section can be found in accompanying demo video\footnote{\url{https://cg.cs.tsinghua.edu.cn/people/~Yongjin/Yongjin.htm}}.

\begin{figure*}[t]
\begin{center}
\includegraphics[width=1.0\textwidth]{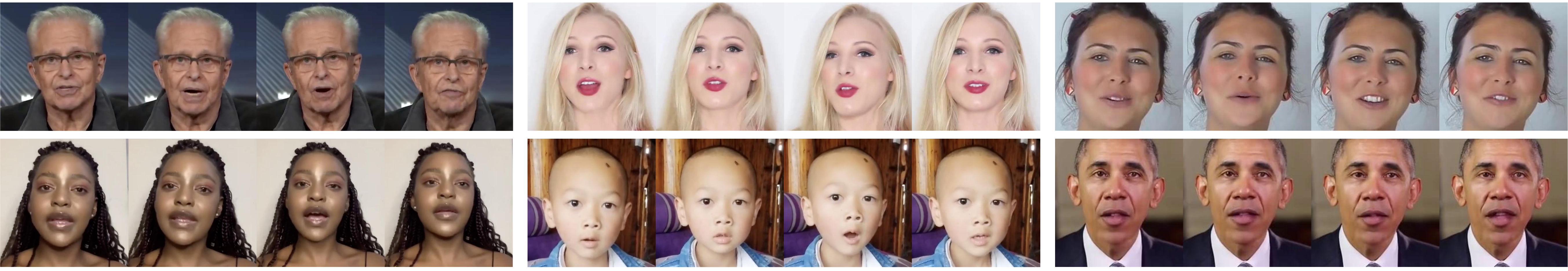}
\end{center}
\caption{Our method works well for people of different races and ages.} 
\label{fig:races_ages}
\end{figure*}

\subsection{Experiment setup}
\label{ssec:setup}
In our model, the two components (audio to expression and pose by LSTM, and memory-augmented GAN) involves two training steps: (1) a general mapping trained by the LRW video dataset~\cite{ChungZ16} and (2) fine tuning the general mapping to learn the personalized talking behavior.
At the fine tuning step, we collect 15 real-world talking face videos of single person from Youtube. In each video, we use its first 12 seconds (about 300 frames) as the training data.
Given the well-trained general mapping, we observe that 300 frames\footnote{See Appendix B for details of this observation.} are sufficient for the fine tuning task. 
In Section \ref{ssec:comparison}, we evaluate our personalized fine tuning effect (see Figure \ref{fig:personalized_rst} for two examples) on both the audio from the original Youtube video and the audio from the VoxCeleb and TCD dataset. Below we denote the general mapping and fine-tuned personalized mapping as Ours-G and Ours-P, respectively. The network is first trained in Ours-G (using general dataset) and then fine-tuned in Ours-P (for a specific person).

In our experiments, the parameters in Eq.(\ref{eq:lstmloss}) is $\lambda_1=0.2$, $\lambda_2=0.01$ and $\lambda_3=0.0001$.
The parameters in Eq.(\ref{eq:ganloss}) is $\lambda_1=100$, $\lambda_2=2$, $\lambda_3=1e-5$.
Our method works well for people of different races and ages. Some examples are illustrated in Figure~\ref{fig:races_ages}.

\begin{figure}[t]
\centering
\includegraphics[width = 0.5\textwidth]{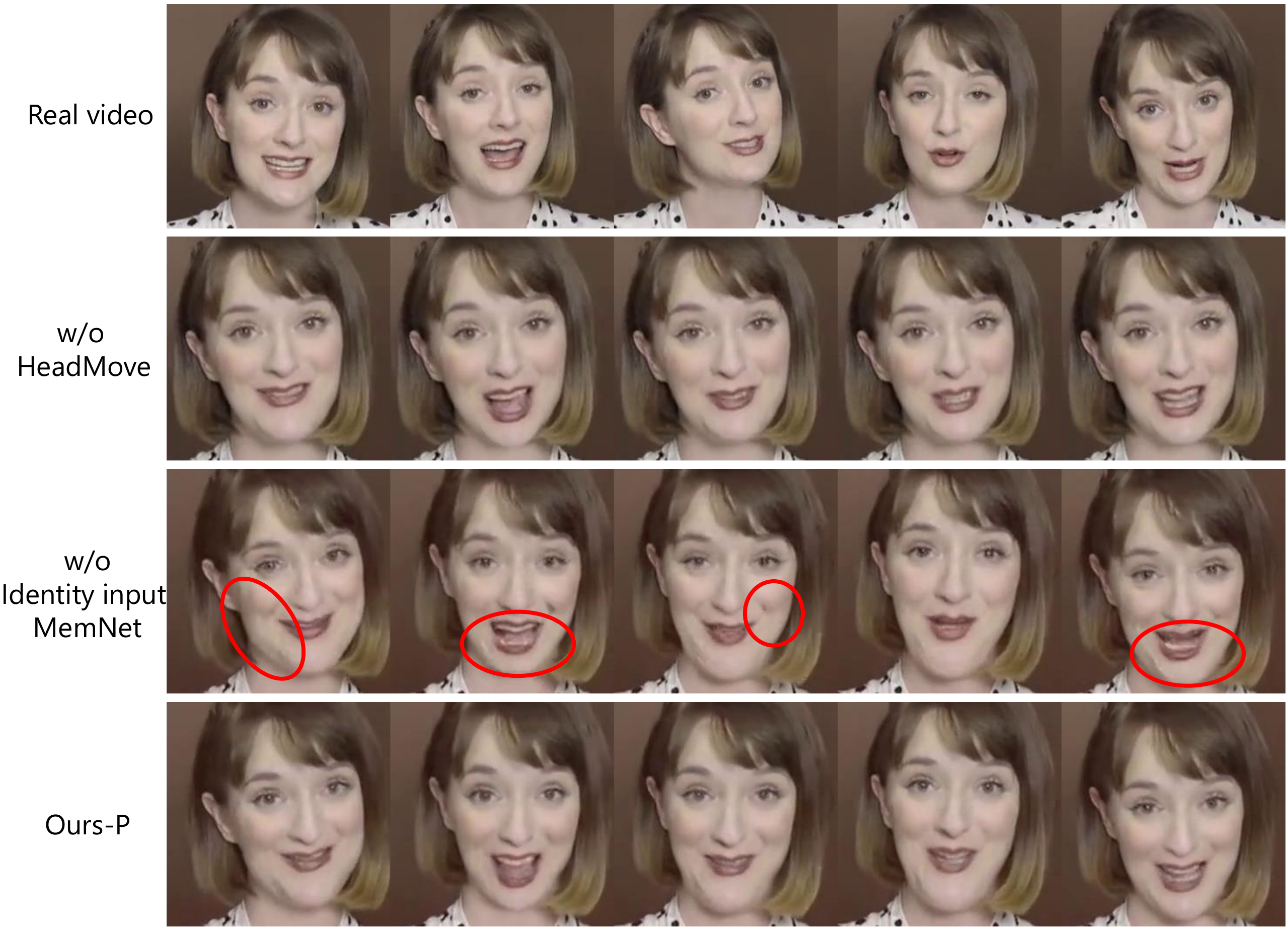}
\caption{Ablation study. The first row shows the ground truth video (a segment from Youtube video). The second row shows the generated results without pose estimation in the first stage. The third row shows the generated results by excluding the identity feature from input of GAN and the memory network from the GAN model. The last low shows the results of our full model.} 
\label{fig:ablation}
\end{figure}

\begin{figure}[t]
\centering
\includegraphics[width = 0.5\textwidth]{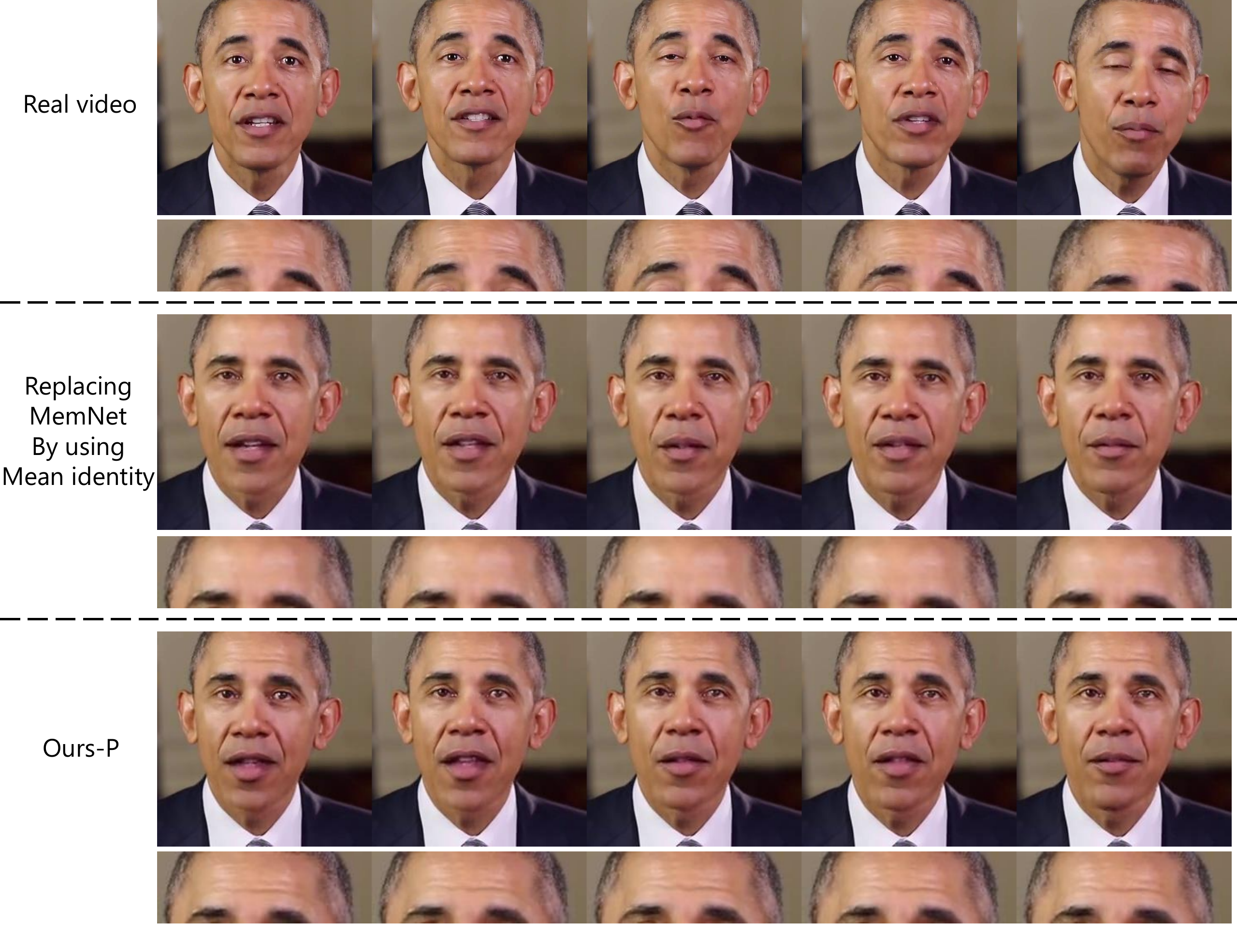}
\caption{Ablation study of replacing the memory network by using the mean of identity vectors in fine tuning. The first row shows the ground truth video (a segment from Youtube video). The middle row shows the results by replacing the memory network with the mean of identity vectors. The last low shows the results of our model.} 
\label{fig:ablation1}
\end{figure}

\subsection{Ablation study}

As illustrated in Figure \ref{fig:pipeline}, our method involves two stages. Here we evaluate the importance of these two stages.

In the first stage, we predict both the head pose and expression from the input audio. If we only predict the expression without the pose estimation, as shown in the second row of Figure \ref{fig:ablation}, the generated results are good in lip synchronization, but look rigid due to the fixed head position, which is far from natural.

There are two distinct characteristics in the second stage. First we include the identity feature in the input of GAN.
Second, we add a memory network in the GAN model to store representative identities in the training set and retrieve the best-matched identity feature during the test.
If we exclude the identity feature from input and the memory network from the GAN model, the personalized refining effect of different identities and expressions would be the same and then the network can not be well optimized. As shown in the third row of Figure \ref{fig:ablation}, without them, the generated results have bad mouth details (e.g., strange teeth), uneven cheek areas, and black spots on face.
If we exclude the memory network from the GAN model but keep the identity feature input, and use the mean of identity features in fine tuning, the results (the middle row in Figure \ref{fig:ablation1}) are not as good as our results (the last row in Figure \ref{fig:ablation1}), which have much better fine details (e.g., wrinkles) and look more realistic.

\subsection{Comparison with state of the arts}
\label{ssec:comparison}

As mentioned in Section \ref{ssec:setup}, our model involves two important mappings: Ours-G and Ours-P.
Note that (1) the inputs to the personalized mapping Ours-P are a short video of 300 frames (to fine-tune) and an audio, and (2) the inputs to the general mapping Ours-G are only one frame (since it does not need to fine-tune) and an audio.

In this section, we show that the Ours-P model can generate realistic talking face video with more distinguishing head movement effects than the state-of-the-art methods. Even for the degenerate case that uses a single face image as input, the Ours-G model can generate comparable lip synchronization with previous methods.

\begin{figure*}[t]
\centering
\includegraphics[width = 0.96\textwidth]{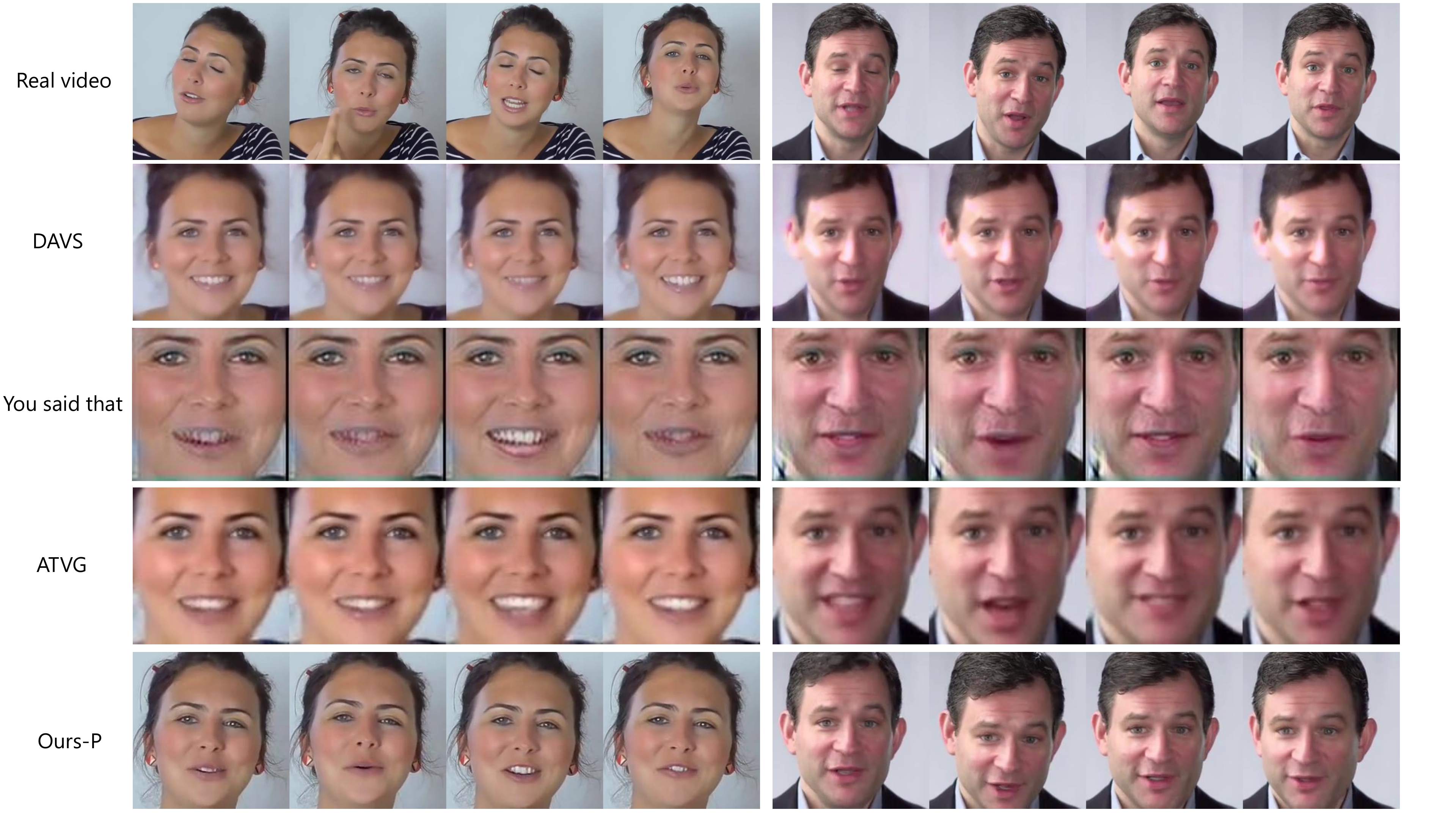}
\caption{Qualitative results of our method and state-of-the-art methods. The first row shows two ground-truth videos (segments from Youtube videos) and the other rows show the results of different methods (DAVS~\cite{ZhouLLLW19}, You said that~\cite{ChungJZ17}, ATVG~\cite{ChenMDX19} and Ours-P). Our method achieves good image quality, lip synchronization (w.r.t. real video) and personalized head movements.}
\label{fig:personalized_compare}
\end{figure*}

\subsubsection{Comparison with Ours-P}

We first compare Ours-P with three state-of-the-art audio-driven talking face generation methods: YouSaidThat~\cite{ChungJZ17}, DAVS~\cite{ZhouLLLW19} and ATVG~\cite{ChenMDX19}.
These three methods are all 2D-based and operate on the image directly, i.e., without using 3D face geometry and rendering. Thus their inputs include only one facial image and an audio. We emphasize that the head positions in the results output from these methods are fixed. Although Ours-P takes more frames (for the purpose of fine-tuning) as input, we learn personalized head pose. Some qualitative results are shown in Figure \ref{fig:personalized_compare}.

It is very challenging to evaluate the visual quality and naturalness of synthesized videos, in particular regarding the human face.
We therefore design a user study to perform the assessment based on subjective score evaluation.
To fine tune the network by considering personalized talking behavior, we collect 15 real-world talking face videos and use the portion of their first 12 seconds for training 15 personalized mappings.
In our user study, for each of these 15 personalized mappings, we test two sets of audio: one is the audio from the remaining portions of the original real videos, and the other is an audio chosen from VoxCeleb~\cite{NagraniCZ17} or TCD~\cite{HarteG15}.
We choose these two datasets because they have a long audio segment to better visualize the change of head pose.
Then we can construct 30 comparison groups.
Each group have five videos: one original video and four generated videos by four methods. Each personalized video devotes to two groups, based on its two sets of audios.

20 participants attended the user study and
each of them compared all 30 groups and answered 3 questions for each group.
For a fair comparison, each group is presented by a randomly shuffled order of five videos.
Participants are asked to select the best video according to three criteria: image quality, lip synchronization and the naturalness of talking.
The results of subjective scores are summarized in Table~\ref{table:user_study}, showing that
our method achieves better performance in all three criteria.

\begin{table}[t]
  \centering
  \caption{Subjective score evaluation of our method and state-of-the-art methods. Each row shows the percentages of a method chosen as the best for different criteria.}
  \setlength\tabcolsep{1pt}
  \begin{tabular}{c|c|c|c}
  \hline
  Methods & Image quality & Lip synchronization & Natural\\
  \hline
  DAVS~\cite{ZhouLLLW19} & 2.17\% & 2.33\% & 2.67\%\\
  \hline
  You said that~\cite{ChungJZ17} & 3.50\% & 20.50\% & 4.17\%\\
  \hline
  ATVG~\cite{ChenMDX19} & 5.67\% & 32.33\% & 9.50\%\\
  \hline
  Ours-P & 88.67\% & 44.83\% & 83.67\%\\
  \hline
  \end{tabular}
  \label{table:user_study}
\end{table}

\begin{table}[t]
  \centering
  \caption{Quantitative results of our method and state-of-the-art methods (Chen~\cite{ChenLMDX18}, Wiles~\cite{WilesKZ18}, You said that~\cite{ChungJZ17}, DAVS~\cite{ZhouLLLW19} and ATVG~\cite{ChenMDX19}), evaluated on LRW dataset which contains 25,000 videos.
  All the methods are evaluated using the same evaluation criterion from ATVG.}
  \setlength\tabcolsep{1pt}
  \begin{tabular}{c|c|c|c|c|c|c}
   \hline
  Methods & Chen & Wiles & You said that & DAVS & ATVG & Ours-G \\
  \hline
  PSNR & 29.65 & 29.82 & 29.91 & 29.90 & 30.91 & 30.94\\
  \hline
  SSIM & 0.73 & 0.75 & 0.77 & 0.73 & 0.81 & 0.75\\
  \hline
  LMD & 1.73 & 1.60 & 1.63 & 1.73 & 1.37 & 1.58\\
  \hline
  \end{tabular}
  \label{table:general}
\end{table}

\subsubsection{Comparison with Ours-G}

Since most previous talking face generation methods do not consider personalized head pose, we further compare our Ours-G (i.e., without fine tuning personalized talking behavior) with representative audio-driven methods~\cite{ChenLMDX18,WilesKZ18,ChungJZ17,ZhouLLLW19,ChenMDX19}.
We directly compare the generated results by different methods with the ground-truth videos.

\begin{figure*}[t]
\centering
\includegraphics[width = 0.98\textwidth]{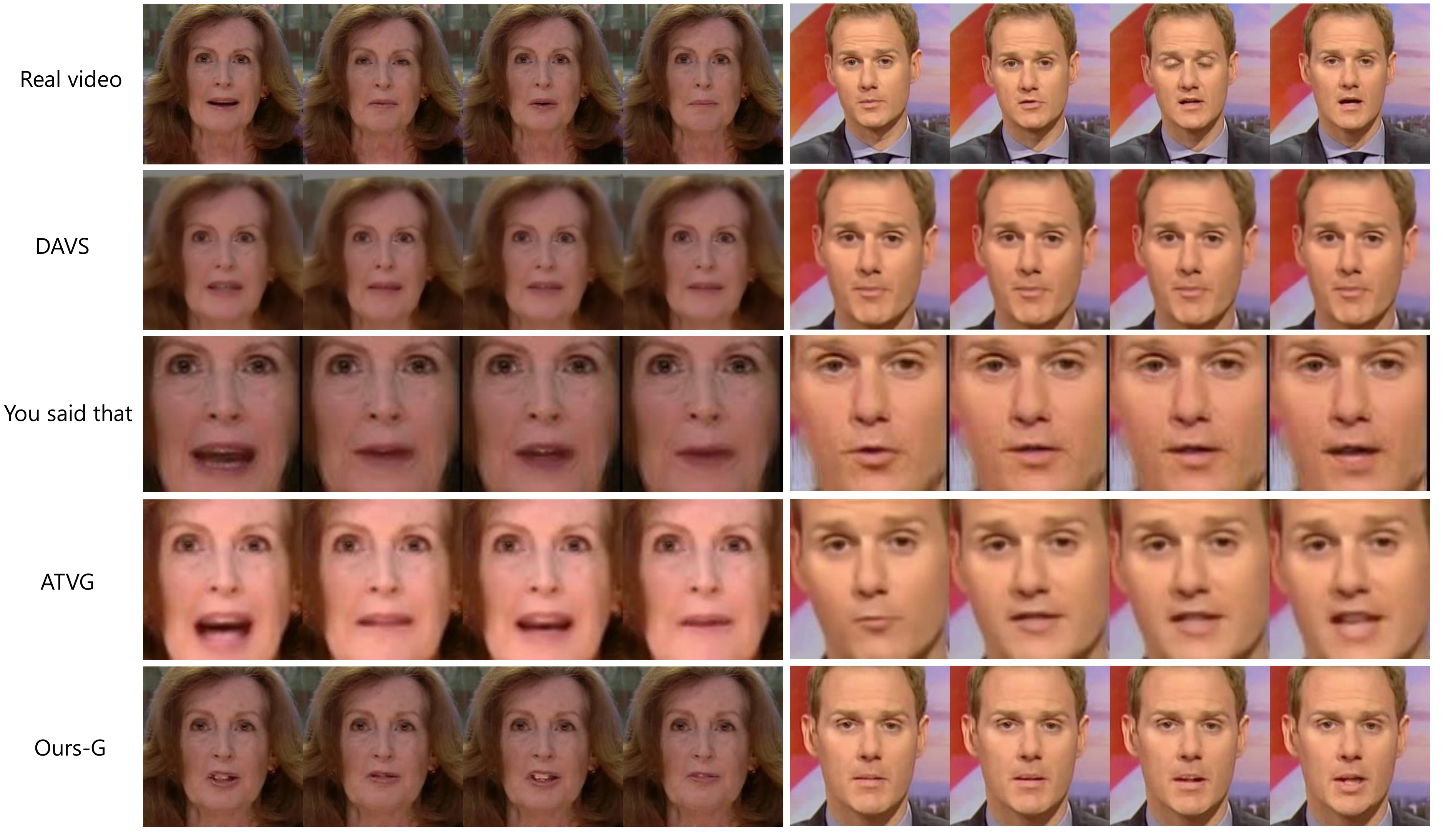}
\caption{Qualitative results of our method (without fine tuning personalized talking behavior) and state-of-the-art methods. The first row is two examples of the ground truth, taken from LRW dataset~\cite{ChungZ16}, which says the word ``absolutely'' (the first four columns) and the word ``abuse'' (the last four columns), respectively. The second to the last rows are the generated results from four different methods. Our method has comparable results (well preserving facial texture and good lip synchronization) with ATVG~\cite{ChenMDX19}.}
\label{fig:general_compare}
\end{figure*}

We follow ATVG~\cite{ChenMDX19} to apply three widely used metrics for audio-driven talking face generation evaluation, i.e.,
the classic PSNR and SSIM metrics for image quality evaluation, and the landmark distance (LMD) for accuracy evaluation of lip movement.
The results are summarized in Table~\ref{table:general}, showing that our method has the best PSNR values and has comparable SSIM and LMD metric values with ATVG~\cite{ChenMDX19}.
Some qualitative comparisons are shown in Figure~\ref{fig:general_compare}.

\subsection{Analysis of head pose behavior}

To objectively evaluate the quality of personalized head pose, we propose a new metric $HS$ to measure the similarity of head poses between the generated video and real video.
We use the three Euler angles to model head movements~\cite{KalioubyR05}, i.e., pitch, yaw, and roll corresponding to the movement of head nod, head shake/turn, and head tilt, respectively.
We compute a histogram $P_{real}$ of pose angles in real personalized video, and a histogram $P_{gen}$ of pose angles in the generated video.
Then we compute the normalized Wasserstein distance $W_1$ \cite{villani2003topics} between $P_{real}$ and $P_{gen}$. The lower the distance, the more similar the two head pose distribution.
Our new metric $HS$ is formulated as
\begin{equation}
HS = 1 - W_1(P_{real},P_{gen})
\end{equation}
where $HS$ is in the range $[0,1]$ and larger $HS$ indicates higher similarity of head pose.
The average $HS$ score of 15 pairs of personalized videos is $0.859$, and the maximum and minimum score are 0.956 and 0.702 respectively, showing that our generated video has a high similarity to real video in term of head movement behavior.

To evaluate the validity of this new metric, we further perform another user study to examine the correlation between the subjective evaluation and our metric values.
20 participants attended this user study. Each participant was asked to compare 15 pairs of generated videos and real videos. They ranked from 1 to 5 based on the head pose similarity of the two videos (1-not similar, 2-maybe not similar, 3-don't know, 4-maybe similar, 5-similar).
The results of votes (in parentheses) are 1 (25), 2 (47), 3 (35), 4 (110) and 5 (83).
The average score is $3.60$, and the percentage of scores 4 and 5 (`maybe similar' and `similar') is $64.3\%$. Only $24.0\%$ ranks are `not similar' or `maybe not similar'.
Furthermore, the correlation coefficient between subjective ranking and the $HS$ metric is $0.65$, demonstrating that our metric has strong positive correlation with human perception.



\section{Conclusion}

In this paper, we propose a deep neural network model that generates a high-quality talking face video of a target person who speaks the audio of a source person.
Since natural talking head poses often cause in-plane and out-of-plane head rotations, to overcome the difficulty of rendering a realistic frames directly from input video to output video, we reconstruct the 3D face and use the 3D facial animation to bridge the gap between audio-visual-driven head pose learning and realistic talking face video generation. The 3D facial animation incorporates personalized head pose and is re-rendered into video frames using a graphic engine. Finally the rendered frames are fine-tuned into realistic ones using a memory-augmented GAN module.
Experiments results and user studies show that our method can generate high-quality talking head video with personalized head pose, and this distinct feature has not been considered in state-of-the-art audio-driven talking face generation methods.

\appendices

\section{Correlation between Audio and Head Pose}
\label{appendix-1}

Our proposed deep network model learns a mapping from audio features to facial expression and head pose.
We infer the head pose from an audio, based on the observation that the speaking style of a person in a short period is often consistent.
In our model, we have two training steps: (1) a general mapping trained by LRW, and (2) fine-tuning step using a short video of the target person as the training data.
The head pose pose estimation of the target person is mainly learned during the fine-tuning step, because LRW includes different person's data and their head movements vary; so we design to learn the head movement behavior from the short video of the target person.

To verify the observation that we can infer the head pose from the audio features,
we conduct a correlation analysis between the audio and pose in these short videos.
We represent the audio using the MFCC feature and represent the pose using the three Eular angles (i.e., pitch, yaw, roll).
For a MFCC feature $s$, we find all MFCC features in the same short video that are in its local neighborhood, and calculate the distance between the neighboring MFCC pair and the distance between the corresponding poses. We calculate the correlation between these two distances using a spherical neighborhood with radius$=0.5*|s|$. The average correlation coefficient of 15 short videos is 0.45, and the maximum and minimum correlation coefficients are 0.58 and 0.24 respectively. These results indicate there exists a positive correlation between the audio and pose.

\section{User Study on the Length of Input Short Video}

\begin{figure}[th]
\centering
\includegraphics[width = 0.5\textwidth]{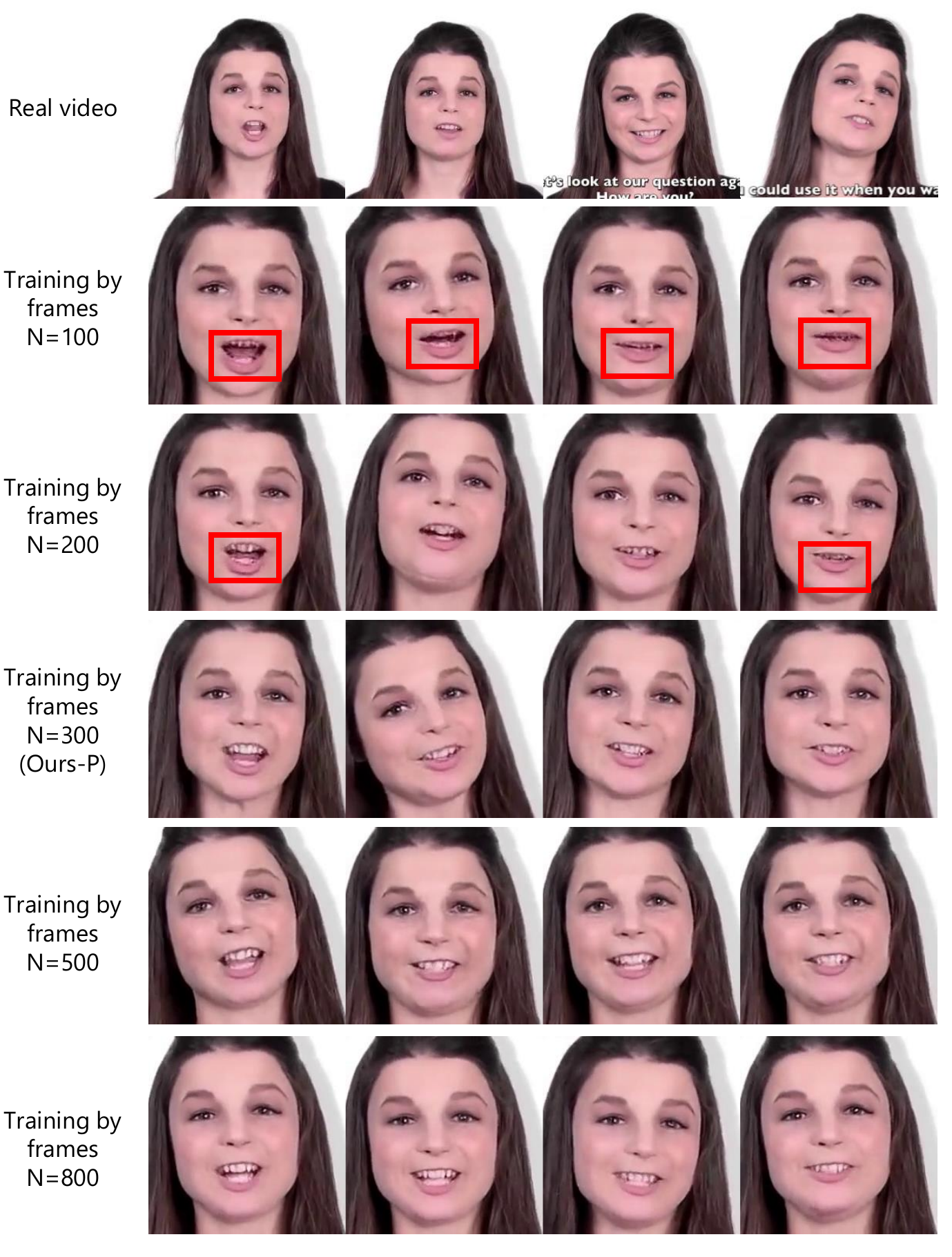}
\caption{Study on the relation between the length of the input short video and the quality of generated results. The first row shows the ground truth video (a segment from Youtube video). The remaining rows show results using the personalized mapping trained by a short video of 100, 200, 300, 500 and 800 frames respectively.}
\label{fig:ablation2}
\end{figure}

In our method, we use a short video of a target person to fine tune the general mapping into a personalized mapping, which learns personalized talking behavior. Here, we study the relation between the length of the input short video and the quality of the output talking face video. We generate the results by inputting short videos of different lengths, i.e., 4s (100 frames), 8s (200 frames), 12s (300 frames), 20s (500 frames), and 32s (800 frames).
Some qualitative results are shown in Figure~\ref{fig:ablation2}.

We first conducted an expert interview and asked an expert who is good at video quality assessment to choose the results that have the best quality and explain why.
The expert chose the results trained by 300, 500 and 800 frames, and the reason was that the results trained with less frames have obviously lower image quality around the mouth and teeth areas, and somehow look strange.

Then we further conducted the following user study.
We asked each user to (1) watch a real video, (2) watch the results generated by the model trained with $N=100, 200, 300, 500, 800$ frames, and (3) select the best ones (in terms of visual quality) from generated results. Note that in our user study, more than one result can be selected as the best; i.e., the user can select multiple results that have the same best quality.
11 participants attended this user study. For the results generated with $N=100, 200, 300, 500, 800$ frames, $0\%, 0\%, 36.4\%, 36.4\%, 63.6\%$ users selected them as the best one, respectively.
These results validated that the models trained by less than 300 frames produce apparently worse results and the model trained by 300 frames achieves a good balance between visual quality and computational efficiency (using fewer frames for training).

{\bf Discussion on frame numbers in personalized and general mapping.} $N=100$ generates low video quality, possibly because in the personalized mapping (i.e. fine-tuning by the input short video), we use the low-frequency albedo (to tolerate the change of head pose), and it requires more frames to fine-tune the low-frequency albedo into realistic ones. While in the general mapping (i.e. trained by LRW, without fine-tuning by the short video), we use the detailed albedo. So using only one frame in the general mapping may generate a little bit more realistic results than using a few frames (e.g., 10-30 frames) to fine-tune the albedo in the personalized mapping.

\ifCLASSOPTIONcaptionsoff
  \newpage
\fi

\bibliographystyle{IEEEtran}
\bibliography{journal}

\end{document}